\documentclass[11pt]{article}

\usepackage[preprint]{acl}

\usepackage{times}
\usepackage{latexsym}
\usepackage{multirow}
\usepackage{enumitem}
\usepackage[T1]{fontenc}

\usepackage[utf8]{inputenc}

\usepackage{microtype}

\usepackage{inconsolata}

\usepackage{graphicx}
\usepackage[table,xcdraw]{xcolor}
\usepackage{adjustbox}

\usepackage{booktabs}

\usepackage[most]{tcolorbox}
\usepackage{fontawesome5}
\usepackage{verbatim}
\usepackage{stfloats}
\usepackage{listings}

\usepackage{hyperref}

\newcounter{promptctr}
\renewcommand{\thepromptctr}{Prompt~\Alph{promptctr}}

\definecolor{examplebg}{HTML}{D1FFBD}
\tcbset{
  bw:domain/.style={
    colback=white,
    colframe=black!40,
    coltitle=black,
    fonttitle=\bfseries\sffamily,   
    colbacktitle=examplebg,
    boxrule=0.5pt,
    titlerule=0pt,
    arc=2pt,
    left=4pt,right=4pt,top=4pt,bottom=4pt,
    toptitle=2pt,bottomtitle=2pt,
  }
}

\newcommand{\mytcbinputwide}[5]{
  \begin{figure*}[t]
  \centering
  \refstepcounter{promptctr}
  \phantomsection
  \begin{tcolorbox}[title={\thepromptctr: #2},#4,width=\textwidth,enhanced]
    \lstinputlisting{#1}
    \label{#5}
  \end{tcolorbox}
  \vspace{-4pt}
  \end{figure*}
}

\newcommand{\promptref}[1]{%
  \hyperref[#1]{\ref*{#1}}%
}

\newcommand{\promptrefp}[1]{%
  \hyperref[#1]{\ref*{#1} (p.~\pageref*{#1})}%
}

\newcommand{\method}{\textsc{TabReX}}
\newcommand{\benchmark}{\textsc{TabReX-Bench}}
\title{\method{}: Tabular Referenceless eXplainable Evaluation}

\author{%
  \textbf{Tejas Anvekar\raisebox{0.75ex}{\includegraphics[height=2ex]{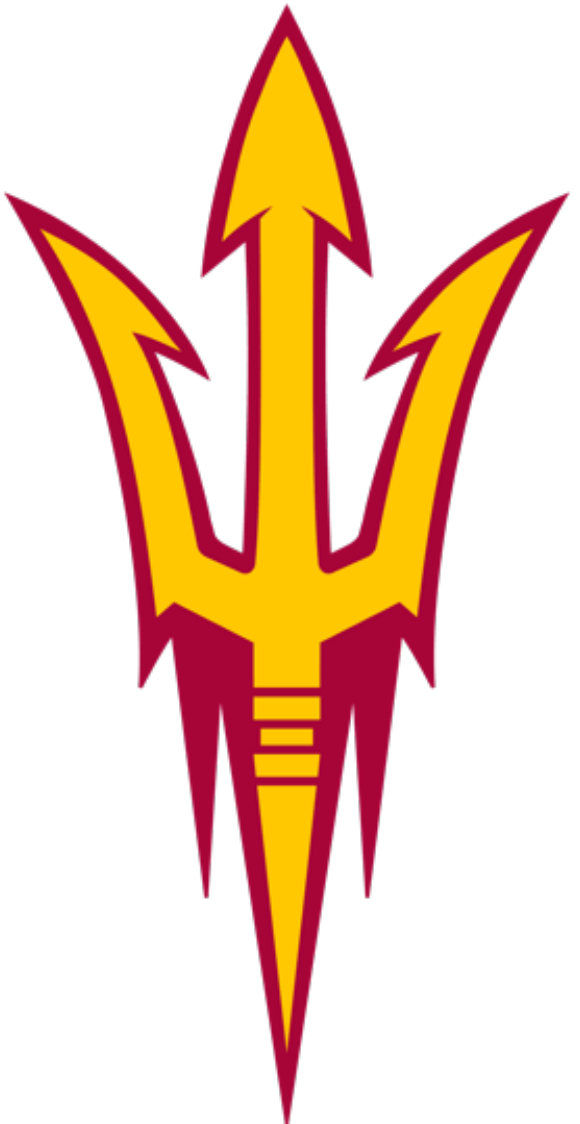}}} \quad
  \textbf{Junha Park\raisebox{0.75ex}{\includegraphics[height=2ex]{./images/asu_logo.pdf}}} \quad
  \textbf{Aparna Garimella\raisebox{0.75ex}{\includegraphics[height=2ex]{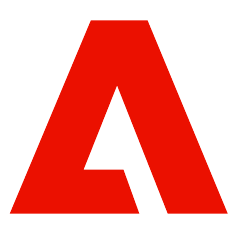}}} \quad
  \textbf{Vivek Gupta\raisebox{0.75ex}{\includegraphics[height=2ex]{./images/asu_logo.pdf}}} \\[0.5em]
  \raisebox{0.75ex}{\includegraphics[height=2ex]{./images/asu_logo.pdf}}\hspace{0.3ex}Arizona State University \quad
  \raisebox{0.75ex}{\includegraphics[height=2ex]{./images/adobe_logo.pdf}}\hspace{0.3ex}Adobe Research \\[0.4em]
  \faGlobe\;\href{https://coral-lab-asu.github.io/TabReX/}{Project-Page} \quad \faGithub\;\href{https://github.com/CoRAL-ASU/TabReX}{Code}\\
  \texttt{\{tanvekar,jpark284,vgupt140\}@asu.edu} \\
  \texttt{garimell@adobe.com}
}

\begin{document}
\maketitle
\begin{abstract}
Evaluating the quality of tables generated by large language models (LLMs) remains an open challenge: existing metrics either flatten tables into text, ignoring structure, or rely on fixed references that limit generalization. We present \textbf{\method}, a \emph{reference-less, property-driven} framework for evaluating tabular generation via graph-based reasoning.  
\method{} converts both source text and generated tables into canonical knowledge graphs, aligns them through an LLM-guided matching process, and computes interpretable, rubric-aware scores that quantify structural and factual fidelity. The resulting metric provides controllable trade-offs between \emph{sensitivity} and \emph{specificity}, yielding human-aligned judgments and cell-level error traces. To systematically assess metric robustness, we introduce \textbf{\benchmark{}}, a large-scale benchmark spanning six domains and twelve planner-driven perturbation types across three difficulty tiers. Empirical results show that \method{} achieves the highest correlation with expert rankings, remains stable under harder perturbations, and enables fine-grained model-vs-prompt analysis establishing a new paradigm for \emph{trustworthy, explainable evaluation} of structured generation systems.
\end{abstract}

\begin{figure}[ht]
    \centering
    \includegraphics[width=\columnwidth]{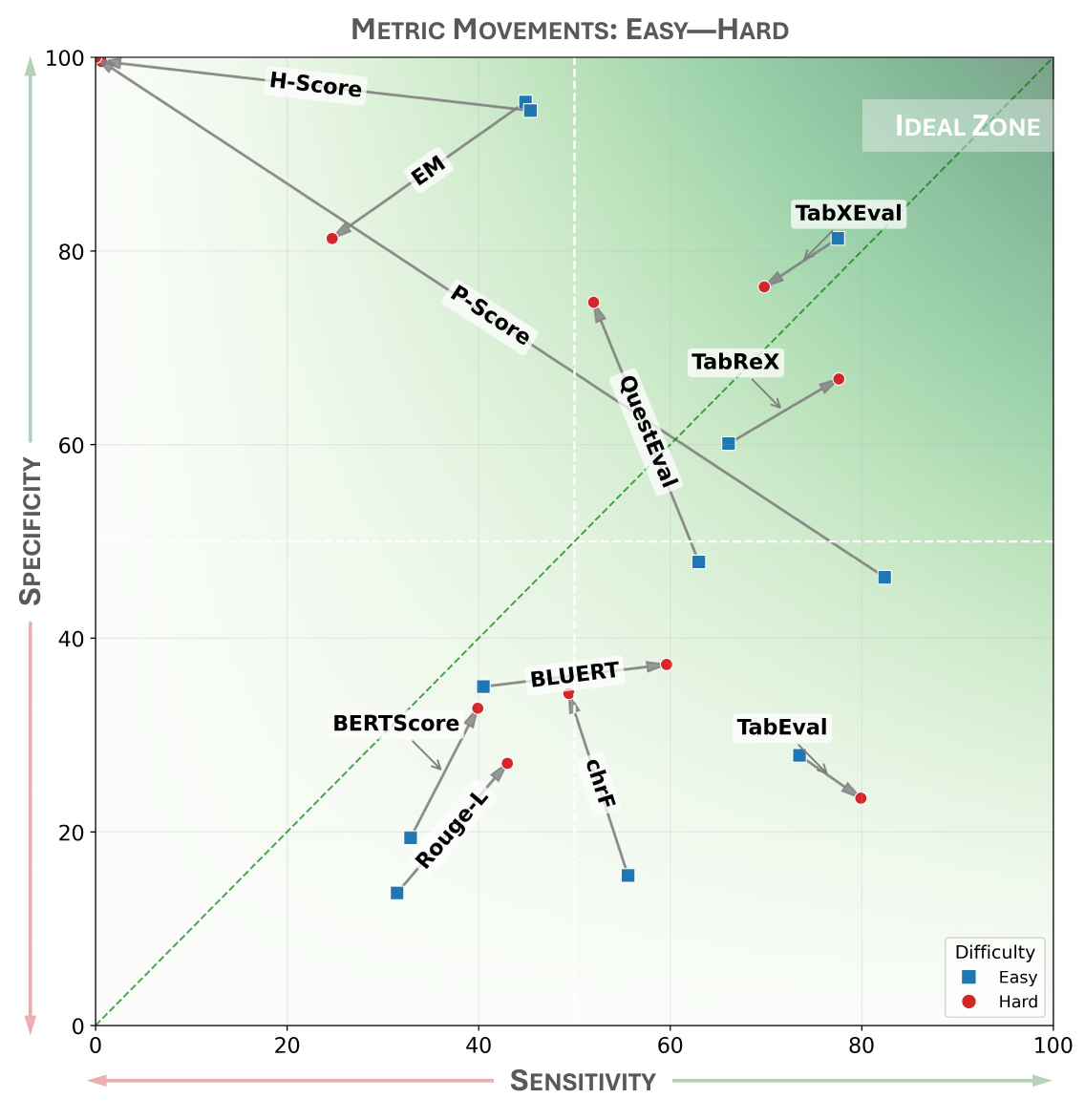} 
    \caption{\textbf{Metric Movements Across Difficulty Levels.} Arrows show each metric’s shift from \textit{easy} (\textcolor{blue}{blue}) to \textit{hard} (\textcolor{red}{red}) perturbations. Axes plot \textit{specificity} (y) vs.\ \textit{sensitivity} (x), with the green region denoting the balanced \textit{ideal zone}. The dashed diagonal marks the optimal trade-off. \method{} stay near this zone, maintaining right direction even for hard examples.}

    \label{fig:metric_movements}
\end{figure}

\section{Introduction}

Structured data underpins critical workflows across domains such as finance, healthcare, scientific reporting, and logistics. Beyond spreadsheets and relational tables, modern ecosystems rely on JSON records, knowledge graphs, and visual dashboards. These formats enable consistent reasoning and aggregation, yet even a single misplaced column, unit mismatch, or corrupted cell can propagate costly downstream errors.

As large language models (LLMs) increasingly generate or transform structured outputs e.g., converting reports into financial tables, synthesizing patient dashboards, or reformatting analytical data the need for \emph{reliable automatic evaluation} has become a major bottleneck. Unlike free-form text, structured generation demands assessment of not just semantic fidelity but also schema alignment, syntactic consistency, and cell-level correctness.

Most existing metrics, however, flatten tables into plain text. N-gram scores like BLEU~\cite{papineni-etal-2002-bleu} and ROUGE~\cite{lin-2004-rouge} ignore row-column structure and unit semantics, while embedding-based metrics such as \textsc{BERTScore}~\cite{bert_score} and \textsc{BLEURT}~\cite{bleurt} capture semantics but miss structural perturbations. Token-level methods like Exact Match or \textsc{PARENT}~\cite{parent} cannot distinguish harmless reformatting from genuine factual errors. Reference-less QA metrics such as \textsc{DataQuestEval}~\cite{dataquesteval} ground evaluation in source evidence but over-penalize layout changes, and recent \textsc{TabEval}~\cite{tabeval} and \textsc{TabXEval}~\cite{tabxeval} improve explainability yet remain limited by small, single-pass benchmarks and one-shot perturbation schemes.

We argue that next-generation evaluation must be both \emph{property-driven} and \emph{personalizable}. Effective metrics should obey key properties-permutation and format invariance, schema- and unit-consistent alignment, monotonic improvement as errors are fixed, and robustness to outliers while allowing tunable trade-offs between \emph{sensitivity} (coverage) and \emph{specificity} (hallucination control). Real-world domains differ in their error tolerance (e.g., precision in finance vs.\ recall in clinical data), requiring metrics that are domain-agnostic by design yet easily adaptable through interpretable property weights.

To meet these needs, we propose \textbf{\method}, a graph-based, explainable evaluation framework. \method{} converts both reference text and generated tables into structured graphs via a hybrid pipeline: a rule-based \emph{Table2Graph} converter and an LLM-assisted \emph{Text2Graph} extractor-followed by an LLM-guided \emph{Graph Alignment} that identifies factual correspondences and discrepancies. From these alignments, a \emph{property-driven scoring} function computes interpretable, rubric-aware penalties capturing both structure and content quality, yielding an explainable, reference-less score.

To stress-test metric reliability, we introduce \textbf{\benchmark}, a large-scale benchmark covering six domains (finance, healthcare, hierarchical tables, and narratives) and twelve planner-driven perturbation types across three difficulty levels. Unlike prior one-shot datasets, \benchmark{} systematically combines factual and structural edits ranging from benign reformatting to severe semantic corruption enabling robust sensitivity-specificity analysis under realistic perturbation regimes.

In summary, our contributions are:
\begin{itemize}[leftmargin=*,itemsep=0.1em]
    \item[-] \textbf{\method}: a \emph{reference-less, property-driven} evaluation framework that aligns table–text graphs and computes interpretable, rubric-aware scores.  
    \item[-] \textbf{\benchmark}: a large, systematically perturbed dataset enabling reproducible metric evaluation across domains and difficulty levels.  
    \item[-] Empirical results showing that \method{} achieves strong human correlation and robustness under harder perturbations.  
    \item[-] Rubric-wise analyses demonstrating that \method{} provides explainable diagnostics at both table and cell levels for model–prompt alignment.
\end{itemize}

\section{\method}
\label{sec:tabrex}

\begin{figure}[!ht]
    \centering
    \includegraphics[width=\columnwidth]{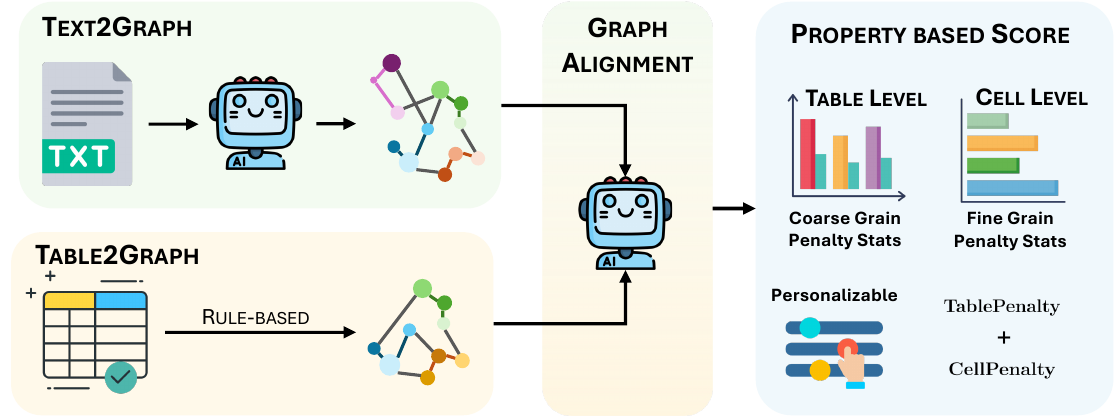} 
     \caption{Illustration of proposed \textbf{\method}. Both source text and generated tables are converted into knowledge graphs via \emph{Text2Graph} and \emph{Table2Graph}, aligned through an LLM-guided \emph{Graph Alignment}, finally scored by a \emph{Property-Driven Scoring} function that aggregates alignment statistics into interpretable, controllable table- and cell-level penalties.}
    \label{fig:main}
\end{figure}

We propose \textbf{\method}, a unified evaluation framework for tabular generation that converts both candidate table and reference / source text into knowledge graphs and scores them through a small set of \emph{property-driven} signals. This design yields a metric that is \emph{reference-less}, \emph{effective in detecting true discrepancies}, and \emph{explainable} by construction, best illustrated in \autoref{fig:main}

\subsection{Pipeline Overview}

\paragraph{Stage 1: Text2Graph and Table2Graph.}  
To enable uniform comparison, \method{} represents both textual summaries and tables as knowledge-graph triplets $[s,p,o]$.

For \textbf{text}, we use an LLM guided by a strict entity-centric grammar (\promptref{prompt:t2g}) to extract minimal atomic facts, where the \emph{subject} is an entity or time slice, the \emph{predicate} a normalized property, and the \emph{object} a canonical value. This design enforces consistent granularity, normalized predicates, and unit-aware values across free-form text: $\mathcal{G}_S=\{(s_i,p_i,o_i)\mid i=1,\dots,n\}.$

For \textbf{tables}, we apply a lightweight \emph{rule-based unrolling}.  
Headers define predicates; each row specifies a subject; every non-empty cell yields a triplet
$(s_{\text{row}},\,p_{\text{header}},\,o_{\text{cell}})$.  
To support diverse table formats, we implemented both \texttt{RuleHTMLConverter} and \texttt{RuleMDConverter}, and in this work, we use the latter. This deterministic approach is fast, schema-aware, and requires no training.

By converting both modalities into a common, interpretable triplet space, \method{} ensures structural clarity and prepares them for downstream alignment and scoring.

\paragraph{Stage 2: Graph Alignment.}  
In our reference-less setup, we align the graph extracted from the \emph{generated table},
$\mathcal{G}_{T}$, with that from the \emph{source text}, $\mathcal{G}_{S}$,
so the table can be judged directly against the textual evidence.

Both graphs consist of factual triplets $(s,\,p,\,o)$.
The alignment, guided by an LLM prompt (\promptref{prompt:alignment}),
maps triplets in $\mathcal{G}_{T}$ to their counterparts in $\mathcal{G}_{S}$.

We adopt a two-step procedure:
(i) a deterministic pass aligns triplets with identical or schema-normalized
subject–predicate pairs;  
(ii) an LLM-assisted refinement aligns the remainder,
resolving paraphrases, abbreviations, and compound attributes (e.g., ``GDP growth (YoY)" $\leftrightarrow$ ``growth\_rate\_2021").

Each matched pair is annotated with a difference vector $\Delta$
recording unit-aware numeric gaps, categorical mismatches,
and whether a fact is missing in the table or extra relative to the source.
The resulting aligned set $\mathcal{A}$ exposes, at the row/column/cell level,
the precise correspondences and discrepancies required for
property-driven scoring.

\paragraph{Stage 3: Property-Driven Scoring.}  
The aligned set $\mathcal{A}$ provides structured evidence of matches, omissions, and numeric deviations between the table graph $\mathcal{G}_{T}$ and the source text graph $\mathcal{G}_{S}$.  
From these alignments, \method{} derives interpretable statistics counts of missing (MI), extra (EI), and partially matched triplets aggregated over rows, columns, and cells. These alignment-derived quantities directly drive two complementary components capturing structural and factual quality.

\begin{equation*}
\label{eq:score-table}
\begin{aligned}
\text{TablePenalty} &=
\beta_\mathrm{MI}
  \Bigl(
    \alpha_r \tfrac{\mathrm{MI}_r}{N_r}
  + \alpha_c \tfrac{\mathrm{MI}_c}{N_c}
  \Bigr)
\\[-1mm]
&\quad+
\beta_\mathrm{EI}
  \Bigl(
    \alpha_r \tfrac{\mathrm{EI}_r}{N_r}
  + \alpha_c \tfrac{\mathrm{EI}_c}{N_c}
  \Bigr),
\end{aligned}
\end{equation*}

\noindent
where $N_r$ and $N_c$ denote the total numbers of rows and columns in $\mathcal{G}_{S}$, and $\mathrm{MI}$ / $\mathrm{EI}$ count missing and extra entities, respectively.  
The \emph{cell-level penalty} captures factual fidelity:

\begin{equation*}
\label{eq:score-cell}
\begin{aligned}
\text{CellPenalty} &=
\beta_\mathrm{MI}\alpha_\mathrm{cell}\tfrac{\mathrm{MI}_\mathrm{cell}}{N_\mathrm{cell}}
+\beta_\mathrm{EI}\alpha_\mathrm{cell}\tfrac{\mathrm{EI}_\mathrm{cell}}{N_\mathrm{cell}}
\\[-1mm]
&\quad+
\beta_\mathrm{partial}\alpha_\mathrm{cell}\tfrac{\Gamma}{N_\mathrm{cell}},
\end{aligned}
\end{equation*}

\noindent
where $\Gamma$ is the sum of normalized numeric deviations over partially aligned cells.  
The final score combines both components:

\begin{equation*}
\label{eq:score-final}
\mathcal{S}_{\text{\method}} = \text{TablePenalty} + \text{CellPenalty}.
\end{equation*}

\noindent
The weighting parameters $(\alpha,\beta)$ provide intuitive control over the metric’s behavior: increasing $\beta_\mathrm{MI}$ favors \emph{sensitivity} (rewarding comprehensive coverage), while increasing $\beta_\mathrm{EI}$ favors \emph{specificity} (penalizing hallucinated entries). Because all quantities are derived directly from $\mathcal{A}$, the score remains reference-less, and fully explainable. All the weight configurations and a walk through example is illustrated in Appendix~\ref{supsec:walkthrough}.

\subsection{\benchmark}
\label{subsec:tabrex-bench}

\begin{figure}[!ht]
    \centering
    \includegraphics[width=0.95\columnwidth]{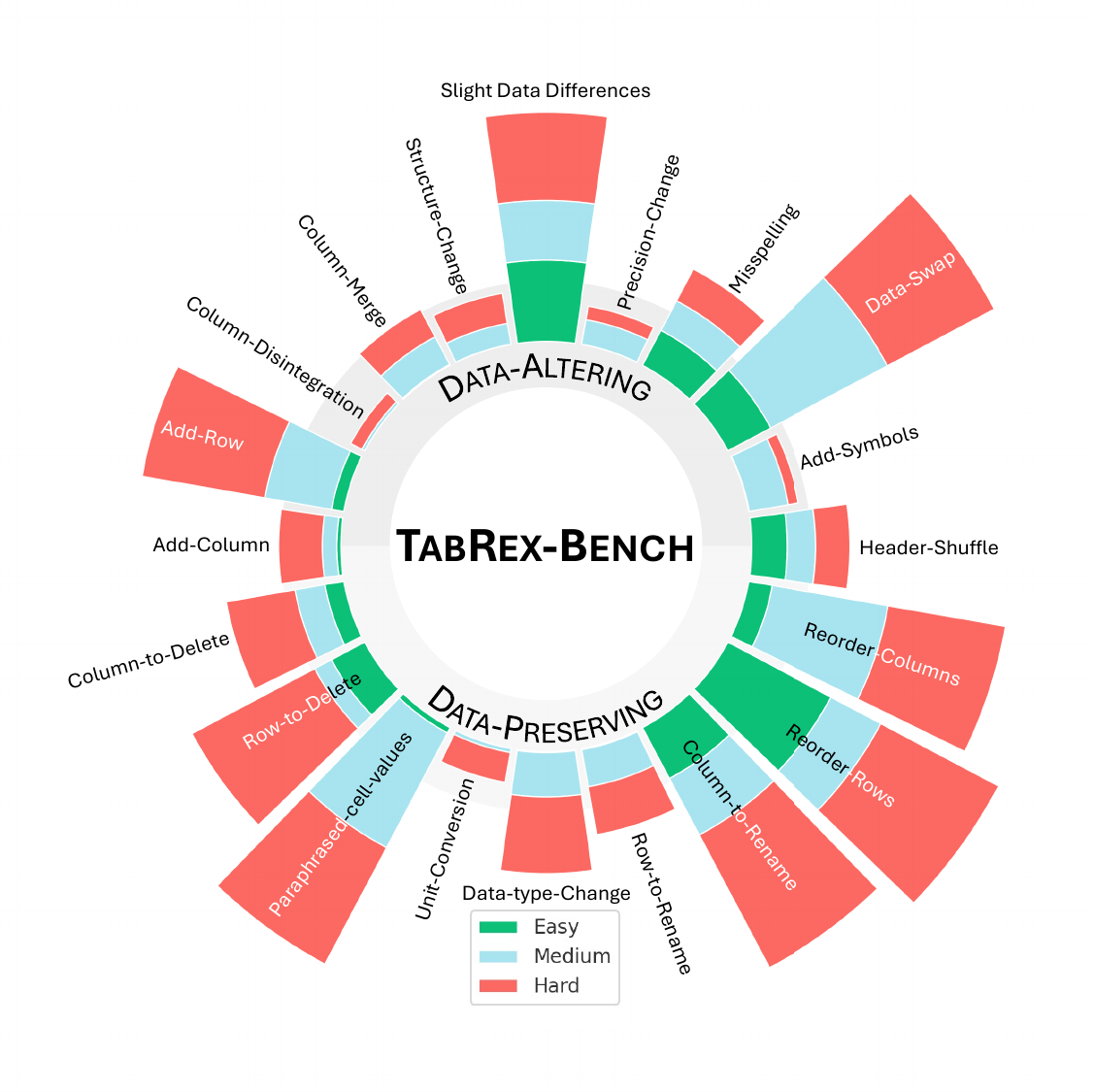} 
    \caption{\textbf{Perturbation landscape across difficulty and type.} 
    The radial stacked donut visualizes the distribution of perturbation types segmented by difficulty: 
    \emph{Easy} (green), \emph{Medium} (blue), and \emph{Hard} (red). 
    The top and bottom semicircles correspond to \emph{data-altering} and \emph{data-preserving} transformations, respectively.}
    \label{fig:tabrexbench-plot}
\end{figure}

\begin{table}[!ht]
\centering
\resizebox{\linewidth}{!}{%
\begin{tabular}{lcccccccc}
\toprule[1.5pt]
\multirow{2}{*}{\textbf{Dataset}} &
\multirow{2}{*}{\textbf{\begin{tabular}[c]{@{}c@{}}\# of \\ Tables\end{tabular}}} &
\multirow{2}{*}{\textbf{\begin{tabular}[c]{@{}c@{}}\# Pert \\ / Table\end{tabular}}} &
\multirow{2}{*}{\textbf{Tables}} &
\multirow{2}{*}{\textbf{\begin{tabular}[c]{@{}c@{}}Avg \\ Row\end{tabular}}} &
\multirow{2}{*}{\textbf{\begin{tabular}[c]{@{}c@{}}Avg\\ Col\end{tabular}}} &
\multirow{2}{*}{\textbf{\begin{tabular}[c]{@{}c@{}}Avg\\ Cell\end{tabular}}} & 
\multirow{2}{*}{\textbf{\begin{tabular}[c]{@{}c@{}}Avg\\ Tokens\end{tabular}}} & 
\multirow{2}{*}{\textbf{\begin{tabular}[c]{@{}c@{}}Avg\\ Num\end{tabular}}} \\
                    &     &    &      &       &       &       &       &       \\ \midrule
\textbf{FinQA}      & 150 & 12 & 1950 & 05.55 & 02.47 & 13.22 & 119.5 & 33.55 \\
\textbf{HiTabQA}    & 150 & 12 & 1950 & 20.08 & 05.60 & 115.1 & 434.8 & 102.7 \\
\textbf{ToTTo}      & 150 & 12 & 1950 & 24.97 & 05.49 & 142.2 & 361.3 & 69.63 \\
\textbf{OpenML med} & 10  & 12 & 120  & 04.20 & 11.58 & 47.94 & 210.9 & 23.80 \\
\textbf{MIMIC-IV}   & 100 & 12 & 1200 & 10.58 & 03.94 & 40.84 & 153.5 & 26.29 \\
\textbf{RotoWire}   & 150 & 12 & 1950 & 10.18 & 05.86 & 59.50 & 146.5 & 14.33 \\ \midrule
\textbf{Total}      & 710 &    & 9120 &       &       &       &       &       \\ \bottomrule[1.5pt]
\end{tabular}%
}
\caption{Statistics of \benchmark: Datasets, perturbation counts, and average table and summary characteristics.}
\label{tab:tabrex_bench_stats}
\end{table}

\benchmark{} is a comprehensive benchmark for evaluating tabular metrics under both \textit{data-preserving} and \textit{data-altering} perturbations. 
Unlike prior resources such as \textsc{TabxBench}~\cite{tabxeval}, which includes only 50 reference tables with 5 perturbations each, 
\benchmark{} spans six heterogeneous datasets FinQA~\cite{finqa}, HiTabQA~\cite{hitabqa}, ToTTo~\cite{totto}, OpenML-med~\cite{openmlmed1,openmlmed7}, MIMIC-IV~\cite{mimic2024}, and RotoWire~\cite{rotowire} covering finance, healthcare, hierarchical tables, and narrative-to-table tasks. As summarized in \autoref{tab:tabrex_bench_stats}, the benchmark comprises $710$ source tables, each expanded with $12$ perturbations, yielding $9{,}120$ perturbed instances spanning compact clinical sheets to large multi-column tables.

\autoref{fig:tabrexbench-plot} illustrates the perturbation composition. 
We define two complementary perturbation groups: \textit{Data-Preserving} (Group~0) alters layout or presentation e.g., row or header reordering, unit conversion, or paraphrasing without changing factual content; \textit{Data-Altering} (Group~1) introduces semantic modifications such as adding or deleting rows/columns, swapping numeric values, or injecting noise and misspellings. Each group is further stratified into three difficulty tiers (\textit{Easy}, \textit{Medium}, \textit{Hard}), supporting controlled analyses of metric robustness as perturbation severity increases.

A key innovation over prior work is our \textbf{planner-driven perturbation generation}. 
Rather than issuing separate LLM calls for each edit, \benchmark{} employs an LLM-based planner (\promptref{prompt:planning}) that generates executable code to produce all $12$ perturbations across both groups and difficulty levels in a single pass, yielding more diverse and reproducible variants. Each perturbed table is also paired with a concise, fact aligned \textbf{table-level summary} (\promptref{prompt:tab2text}) and stats for the avg $\#$ token and Numerical data present are given in \autoref{tab:tabrex_bench_stats}, enabling the evaluation of reference-less metrics assessing factual consistency between tables and summaries an aspect not present in \textsc{TabxBench}.

All perturbations and summaries were initially generated through this planner-driven pipeline and validated on $20\%$ of the data, achieving inter-annotator agreement of $87\%$ for summaries and $91\%$ for perturbations, ensuring correctness and diversity. 
By combining broad domain coverage, structured perturbation design, paired summaries, and tiered difficulty, \benchmark{} enables rigorous evaluation of metric robustness, sensitivity, and human alignment across both reference-based and reference-less settings.

\section{Experiments}
To assess the efficacy of \method, we conduct experiments using our synthetic benchmark \benchmark. All results are reported with \texttt{GPT-5-nano}~\cite{gpt5}, evaluating both components of \method: \textit{Text2Graph} and \textit{Graph Alignment} using proposed \benchmark{} dataset. \\

\paragraph{Baselines.}
We compare \method{} against a diverse set of automatic evaluation metrics grouped by methodological design.
\textit{Deterministic} metrics: Exact Match (\textsc{EM}), \textsc{chrF}, and \textsc{ROUGE-L}: compute token- or character-level overlaps, offering reproducible yet surface-biased comparisons.
\textit{Algorithmic} metrics such as \textsc{H-Score} perform structured alignment and rule-based matching without relying on neural embeddings, offering deterministic, training-free evaluation.
\textit{Neural} metrics such as \textsc{BERTScore} and \textsc{BLUERT} leverage contextual embeddings to capture semantic similarity but may exhibit variability across runs.
Among recent LLM-based approaches, we include \textsc{P-Score} (an LLM-judged quality metric producing 0–10 scores) and \textsc{TabEval}, which flattens tables via an LLM and measures entailment using RoBERTa-MNLI.
We also evaluate the state-of-the-art \textsc{TabXEval}, a two-phase rubric-based framework that first aligns tables structurally (\textit{TabAlign}) and then performs semantic and syntactic comparison (\textit{TabCompare}) for interpretable, human-aligned evaluation.
Finally, we benchmark the reference-less \textsc{QuestEval}, which generates question–answer pairs from both the source and the generated text or table, performs cross-validation using two LLM calls, and computes F1 scores to measure factual and semantic consistency. \\

\paragraph{LLMs.}  
We conduct all experiments using \texttt{GPT-5-nano}, \texttt{Gemma-3}~(4B/27B-Instruct)~\cite{gemma}, and \texttt{InternVL3.5}~(8B-Instruct/Thinking)~\cite{internvl}. Unless stated otherwise, we employ uniform decoding settings across models, using their default temperature, top-$k$, and top-$p$ parameters. All gpu-intensive experiments were conducted on NVIDIA-2$\times$H100s. The full prompts for \textit{Text2Graph} (\promptref{prompt:t2g}) and \textit{Graph Alignment} (\promptref{prompt:alignment}) are provided in Appendix~\ref{supsec:prompts}.

\subsection{Correlation Analysis of Metrics Category.}
\begin{table}[!ht]
\centering
\begin{adjustbox}{width=\linewidth}
\begin{tabular}{lcccccc}
\toprule[1.5pt]
\textbf{Metric} & $\rho_{S}$ $\uparrow$ & $\tau_{K}$ $\uparrow$ & $\tau_{w}$ $\uparrow$ & \textbf{RBO} $\uparrow$ & $\zeta_{F}$ $\downarrow$ & $\pi_{t}$ $\downarrow$\\ \midrule
\multicolumn{7}{c}{\cellcolor[HTML]{FCFBD7}\textit{Non-LLM Based (w/ Ref)}} \\
\textbf{\textsc{EM}}        & 45.88 & 39.38 & 39.51 & 43.33 & 47.49 & 58.40\\
\textbf{\textsc{chrF}}      & 41.76 & 34.55 & 31.61 & 39.39 & 49.26 & 01.64\\
\textbf{\textsc{ROUGE-L}}   & 31.18 & 26.69 & 22.56 & 37.65 & 55.94 & 01.97\\
\textbf{\textsc{BLUERT}}    & 44.66 & 37.64 & 36.09 & 39.57 & 48.09 & 00.77\\
\textbf{\textsc{BERTScore}} & 36.21 & 30.66 & 27.96 & 38.11 & 53.25 & 00.92\\
\textbf{\textsc{H-Score}}   & 56.87 & 47.97 & 51.73 & 41.11 & 40.02 & 00.99\\
\multicolumn{7}{c}{\cellcolor[HTML]{D7FFFE}\textit{LLM-Based (w/ Ref)}} \\
\textbf{\textsc{P-Score}}   & 49.24 & 40.00 & 37.43 & 40.73 & 43.93 & 07.39\\
\textbf{\textsc{TabEval}}   & 49.01 & 39.22 & 34.21 & 41.11 & 43.06 & 00.63\\
\textbf{\textsc{TabXEval}}  & \textbf{80.27} & \textbf{72.37} & \textbf{66.87} & \textbf{47.54} & \textbf{20.94} & 45.33\\
\multicolumn{7}{c}{\cellcolor[HTML]{DEFCDE}\textit{(w/o Ref)}} \\
\textbf{\textsc{QuestEval}} & 62.93 & 52.29 & 51.71 & 42.70 & 35.04 & 03.03\\
\textbf{\method}      & \textbf{74.51} & \textbf{64.24} & \textbf{62.28} & \textbf{44.85} & \textbf{27.01} & 13.59\\ \bottomrule[1.5pt]
\end{tabular}
\end{adjustbox}
\caption{
Correlation of automatic evaluation metrics with human rankings across synthetic perturbation sets. Higher values of Spearman’s rank correlation ($\rho_{S}$), Kendall’s tau ($\tau_{K}$), weighted Kendall’s tau ($\tau_{w}$), and Rank-Biased Overlap (RBO) indicate stronger monotonic and positional agreement with human orderings ($\uparrow$), while lower values of Spearman’s footrule distance ($\zeta_{F}$) and tie ratio ($\pi_{t}$) denote better rank stability and finer discriminative resolution ($\downarrow$). The proposed \method{} achieves the best overall consistency with human judgment.}
\label{tab:synthetic_ranking_tab}
\end{table}

\autoref{tab:synthetic_ranking_tab} reports the correlation between automatic evaluation metrics and human judgments over the synthetic perturbation benchmark. 
Each ground-truth (\textsc{GT}) table was paired with twelve systematically perturbed variants six preserving factual content (labels 0: \textit{1-easy}, \textit{1-medium}, \textit{1-hard}) and six introducing data alterations (labels 1: \textit{1-easy}, \textit{1-medium}, \textit{1-hard}). Human annotators ranked these variants by perceived semantic and factual fidelity to the \textsc{GT}, providing a gold human order for correlation analysis. Metrics are grouped by family Non-LLM, LLM-based, and reference-less to examine their consistency and robustness under controlled perturbations.

\paragraph{(a) Non-LLM metrics.} such as \textsc{EM}, \textsc{chrF}, and \textsc{ROUGE-L} show limited alignment with human judgment. Their Spearman’s ($\rho_{S}$) and Kendall’s ($\tau_{K}$) values remain low ($\rho_{S}\!<\!0.45$, $\tau_{K}\!<\!0.35$), indicating that rank orderings diverge substantially from human perception. Sentence-level embedding metrics (\textsc{BLEURT}, \textsc{BERTScore}) capture partial semantic similarity but exhibit modest RBO ($\approx$0.39) and high footrule distances ($\zeta_{F}\!\approx\!45$$53$), reflecting poor rank stability. Their near-zero tie ratios ($\pi_{t}\!<\!2\%$) further suggest coarse differentiation, failing to separate semantically close variants.

\paragraph{(b) LLM-based metrics.} such as \textsc{P-Score}, \textsc{TabEval}, and \textsc{TabXEval} show notably higher agreement with human preferences ($\rho_{S}\!\approx\!0.49$$0.80$, $\tau_{K}\!\approx\!0.39$$0.72$). Among them, \textsc{TabXEval} achieves the strongest overall correlation ($\rho_{S}\!=\!0.80$, $\tau_{K}\!=\!0.72$), confirming that instruction-tuned evaluators capture perturbation sensitivity effectively. However, its elevated tie ratio ($\pi_{t}\!=\!45.3\%$) and moderate rank dispersion ($\zeta_{F}\!=\!20.9$) indicate frequent scoring saturation, where distinct variants receive identical judgments reducing discriminative precision even when global trends align.

\paragraph{(c) Reference-less metrics.} Without access to reference tables, \textsc{QuestEval} maintains moderate alignment ($\rho_{S}\!=\!0.63$, $\tau_{K}\!=\!0.52$) by generating QA pairs from both the source and system outputs, yet exhibits instability under data-altering perturbations. 
In contrast, \textbf{our metric} achieves the most balanced performance across all dimensions Spearman’s $\rho_{S}\!=\!0.75$, Kendall’s $\tau_{K}\!=\!0.64$, and weighted $\tau_{w}\!=\!0.62$ while also maintaining competitive RBO ($44.9$) and low rank dispersion ($\zeta_{F}\!=\!27.0$). Its moderate tie ratio ($\pi_{t}\!=\!13.6\%$) indicates finer discriminative granularity, avoiding overconfidence and reflecting human-perceived difficulty progression. Together, these findings highlight that our method preserves ordinal consistency across perturbation severity while generalizing robustly in the absence of reference data.

\begin{table}[!ht]
\centering
\begin{adjustbox}{width=\linewidth}
\begin{tabular}{lcccccc}
\toprule[1.5pt]
\textbf{Metric} & $\rho_{S}$ $\uparrow$ & $\tau_{K}$ $\uparrow$ & $\tau_{w}$ $\uparrow$ & \textbf{RBO} $\uparrow$ & $\zeta_{F}$ $\downarrow$ & $\pi_{t}$ $\downarrow$\\ \midrule
\multicolumn{7}{c}{\cellcolor[HTML]{F1F8E9}\textit{Ensemble Baselines}} \\
\textbf{Lex-Emb ($M$)}      & 38.43 & 32.65 & 30.17 & 38.52 & 52.15 & 00.49\\
\textbf{Lex-Emb ($H$)}  & 29.80 & 24.00 & 19.68 & 37.65 & 55.04 & 00.63\\
\textbf{LLM ($M$)}          & 48.49 & 39.21 & 36.94 & 40.56 & 44.38 & 00.42\\
\textbf{LLM ($H$)}      & 56.00 & 46.93 & 50.64 & 40.95 & 40.63 & 00.42\\
\textbf{Hybrid ($M$)}       & 32.04 & 24.94 & 20.29 & 37.03 & 51.51 & 01.13\\
\textbf{Hybrid ($H$)}   & 54.03 & 42.71 & 32.61 & 42.31 & 40.11 & 01.13\\ \midrule
\textbf{\method}      & \textbf{74.51} & \textbf{64.24} & \textbf{62.28} & \textbf{44.85} & \textbf{27.01} & 13.59\\ \bottomrule[1.5pt]
\end{tabular}
\end{adjustbox}
\caption{Comparison of ensemble baselines with the proposed \method{}. 
Ensembles combine metric families: Lex-Emb (lexical + embedding), LLM (LLM-based), and Hybrid (reference + reference-less) using either simple Mean (M) or Harmonic (H) aggregation. All ensemble variants fall short of \method, which achieves the highest correlation with human rankings and better rank stability.}
\label{tab:ensemble_comparison}
\end{table}

\paragraph{(d) Ensemble of Scores.}  
We further benchmarked ensemble baselines that aggregate complementary metrics using either simple averaging (\textit{Mean}) or harmonic averaging (\textit{Harmonic}). These ensembles span three families: \textit{Lex-Emb} (\textsc{EM}, \textsc{ROUGE-L}, \textsc{BERTScore}, \textsc{BLEURT}, \textsc{chrF}), \textit{LLM} (\textsc{P-Score}, \textsc{H-Score}), and \textit{Hybrid} (\textsc{TabXEval}, \textsc{QuestEval}). While the best-performing ensemble, \textit{LLM (Harmonic)}, achieves $\rho_{S}=0.56$ and $\tau_{K}=0.47$, it still lags behind our \method, which attains $\rho_{S}=0.75$ and $\tau_{K}=0.64$ with lower rank dispersion. This highlights that naive aggregation of diverse metrics cannot match the targeted, reference-less reasoning of \method{}, which better aligns with human judgment across perturbation severities.

\subsection{Can \method{} Generalize Across Perturbation Regimes?}

A robust evaluation metric must remain reliable not only in standard (\textit{easy}) settings but also under \textit{hard perturbations} tables with subtle misalignments, semantic shifts, or fine-grained numeric errors. Using our proposed \benchmark, we sample both \textit{easy} and \textit{hard} cases across data-preserving and data-changing perturbations to compute true-positive and true-negative rates (sensitivity and specificity). \autoref{fig:metric_movements} plots each metric’s \textit{trajectory} on the specificity–sensitivity plane as difficulty increases, revealing whether it remains stable or degrades under stress.

\paragraph{Embedding-Driven Metrics.}  
Many popular metrics (e.g., \textsc{BERTScore}, \textsc{BLUERT}, \textsc{TabEval}) rely on neural embeddings rather than surface-level string matching. For example, \textsc{TabEval} first unrolls tables into natural-language atomic statements using an LLM, then applies RoBERTa-MNLI~\cite{roberta} to score entailment between candidate and reference statements.  
Such embedding-based approaches capture deeper semantics, yet as \autoref{fig:metric_movements} shows, they still exhibit large drops in sensitivity or specificity under harder perturbations.

\paragraph{Stability vs.\ Fragility.}  
Metrics with only \textit{short arrow movements} from easy to hard cases (e.g., \textsc{TabXEval}, \method) demonstrate stable trade-offs and thus \textit{robust generalization}. Interestingly, even though \textsc{TabXEval} sits in the ideal zone, its trajectory drifts slightly \textit{away} from the optimal direction as difficulty rises.  By contrast, metrics such as \textsc{EM}, \textsc{H-Score}, and even the LLM-based \textsc{P-Score} experience \textit{sharp drops in sensitivity}, revealing an over-reliance on surface-level cues-showing that an LLM backbone alone does not guarantee proper alignment.

\paragraph{Reference-less Metrics.}  
Both \textsc{QuestEval} and our proposed \method{} evaluate tables without explicit references, instead judging how well a candidate table supports automatically generated questions. \textsc{QuestEval} employs an LLM for question generation and a QA module to assess semantic fidelity, but its reliance on generic QA signals often penalizes harmless re-orderings or formatting changes. In contrast, \method{} tailors question generation to tabular structure and integrates explicit reasoning over extracted facts, enabling it to better separate meaningful discrepancies from superficial variations.  
As shown in \autoref{fig:metric_movements}, this specialization helps \method{} stay closer to the ideal zone even under tougher perturbations, reflecting stronger alignment with human judgment.

\paragraph{Towards Trustworthy Evaluation.}  
These results highlight the importance of \textit{balanced, difficulty-robust metrics} for downstream evaluation.  
As generative table models encounter noisier, real-world data, reliable metrics must \textit{reward genuine comprehension} rather than superficial matches. The ability of \method{} to remain in the green “ideal zone’’ across difficulty levels-despite being \textit{reference-less} underscores its suitability for \textit{high-stakes domains} such as scientific reporting and financial auditing, where both \textit{false alarms} and \textit{missed discrepancies} can be costly.

\subsection{Evaluation on Text-to-Table Task}
\label{subsec:realworld_text2table}

To assess robustness of our method in realistic reference-less settings, we evaluate its performance on text-to-table generation across diverse domains including finance, healthcare, and sports.  
Generated tables are produced by strong open and proprietary LLMs (\texttt{Gemma-3-(4/27B)}, and \texttt{InternVL-3.5-thinking (on/off)}).  
Humans ranking generated tables across models and prompting strategies (zero-shot, CoT, Map\&Make).

Expert annotators ranked the model outputs along three axes \emph{structural correctness}, \emph{factual fidelity}, and \emph{semantic coverage}.  
We then measured how well automatic metrics correlate with these human rankings (detailed in Appendix~\ref{supsec:human_eval}) using Spearman’s~$\rho_{S}$, Kendall’s~$\tau_K$, and rank-biased overlap (RBO).

\begin{table}[!ht]
\small
\centering
\caption{Correlation of automatic metrics with human rankings on real-world text-to-table generation. \method{} achieves the highest alignment across all correlation metrics.}
\label{tab:realworld_text2table}
\begin{tabular}{lccc}
\toprule[1.5pt]
\textbf{Metric} & $\rho_{S} \uparrow$ & $\tau_{K} \uparrow$ & RBO $\uparrow$ \\ \midrule
EM & $-0.01$ & $0.01$ & $0.33$ \\
ROUGE-L & $0.33$ & $0.25$ & $0.29$ \\
BERTScore & $0.26$ & $0.19$ & $0.38$ \\
BLEURT & $0.29$ & $0.20$ & $0.39$ \\
CHRF & $0.25$ & $0.19$ & $0.36$ \\
QuestEval (ref-less) & $0.28$ & $0.20$ & $0.39$ \\
TabEval (ref-based) & $0.25$ & $0.19$ & $0.36$ \\
TabXEval (ref-based) & $0.24$ & $0.17$ & $0.37$ \\
\textbf{\method{} (ref-less)} & $\mathbf{0.39}$ & $\mathbf{0.30}$ & $\mathbf{0.41}$ \\ 
\bottomrule[1.5pt]
\end{tabular}
\end{table}

\paragraph{Observations.}
Surface- and embedding-based metrics (e.g., ROUGE-L, BERTScore, BLEURT) exhibit weak correlation with human preferences, primarily due to their sensitivity to lexical and formatting variation.  
QuestEval performs better but remains brittle to domain-specific structure shifts such as nested headers or missing subtables.  In contrast, \method{} achieves the strongest correlations across all measures \textbf{Spearman’s~$\rho=0.39$}, \textbf{Kendall’s~$\tau_b=0.30$}, and \textbf{RBO=0.41} demonstrating superior alignment with expert judgments.  
Its graph-based reasoning captures factual and structural consistency more effectively, validating its reliability as a \emph{reference-less} evaluator for real-world table generation systems.

\subsection{Rubric-wise Model–Prompt Alignment}
\label{subsec:rubricwise_alignment}

\begin{figure*}[t]
    \centering
    \includegraphics[width=\linewidth]{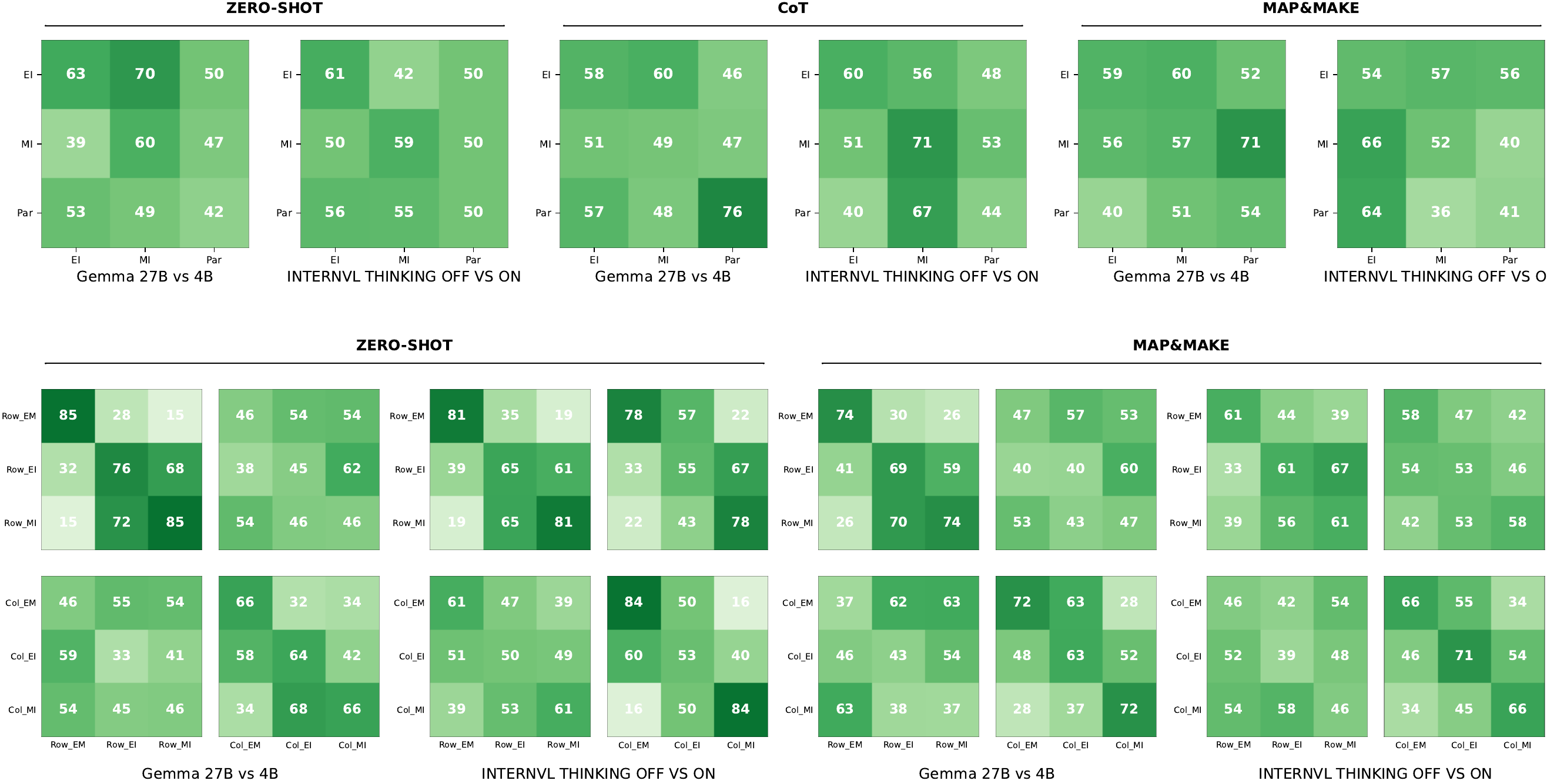}
    \caption{\textbf{Rubric-wise alignment across models and prompting strategies.} 
    Top row: cell-level agreement within model across prompts. 
    Bottom row: table-level agreement. 
    Model size and reasoning style influence local precision more than structural coherence, while prompt strategy (like Map\&Make~\cite{mapandmake}) drives balanced alignment across rubric dimensions.}
    \label{fig:rubricwise_model_alignment}
\end{figure*}

\method{} rubric-aware scoring enables coarse to fine-grained comparison across \emph{models} (e.g., Gemma~8B vs.~27B, InternVL-Thinking On vs.~Off) and \emph{prompting strategies} (Zero-Shot, Chain-of-Thought, Map\&Make~\cite{mapandmake}), measured at both \emph{cell-level} and \emph{table-level} granularity (\autoref{fig:rubricwise_model_alignment}).

\paragraph{Cell-level alignment (top row).} 
Larger models (e.g., Gemma~27B) show clear gains in local fidelity especially for numeric and structural rubrics but only modest improvement in semantic consistency. Reasoning-oriented (``Thinking”) variants improve precision on numeric and structural dimensions yet often underperform on partial or contextual agreement, suggesting over-cautious reasoning can reduce semantic coverage. Chain-of-Thought prompting enhances numeric correctness but sometimes amplifies inconsistency, while Map\&Make maintains more balanced yet slightly conservative performance.

\paragraph{Table-level alignment (bottom row).} 
At a global scale, model size yields diminishing returns: Gemma~27B’s advantage narrows, and ``Thinking” variants do not consistently outperform standard modes. Zero-shot improves row-column coherence but increases rubric variance. Map\&Make achieves steadier rubric alignment, indicating stronger integration of local reasoning into structural organization.

\paragraph{Insights.}
Overall, three trends emerge: (1) larger models enhance fine-grained (cell-level) fidelity but not global coherence; (2) ``Thinking” reasoning improves precision but limits coverage, favoring accuracy over breadth; and (3) prompt design particularly Map\&Make contributes as much as model scale to balanced rubric alignment.

These results illustrate how a referenceless, explainable evaluation metric can reveal the strengths and weaknesses of models and prompting strategies across hierarchical levels. Such rubric-aware scorers enable targeted analysis and can support verifiable reward modeling~\cite{rlvr} for improved alignment.

\section{Comparison with Related Work}
\label{sec:related}

\paragraph{From Text-to-Table to Structural Benchmarks.}  
Early text-to-table datasets such as \textsc{RotoWire} for basketball summaries~\cite{rotowire}, \textsc{E2E} for restaurant descriptions~\cite{novikova-etal-2017-e2e}, \textsc{WikiBio} for infobox biographies~\cite{lebret2016neuraltextgenerationstructured}, and \textsc{WikiTableText}~\cite{wikitables} provided important initial testbeds but offered limited schema diversity and often encouraged hallucinated or under-structured outputs.  Recent resources, including \textsc{StructBench}~\cite{2025structext} and \textsc{TanQ}~\cite{tanq}, introduced challenging phenomena such as header permutations, schema reshuffling, and multi-hop reasoning.  These benchmarks exposed fundamental weaknesses in both generation models and evaluation metrics, motivating the need for metrics that go beyond surface overlap and can reason about structural and semantic fidelity.

\paragraph{Metric Families: From Overlap to Explainability.}  
Conventional reference-based metrics: \mbox{BLEU}~\cite{papineni-etal-2002-bleu}, \mbox{ROUGE-L}~\cite{lin-2004-rouge}, \mbox{METEOR}~\cite{banerjee-lavie-2005-meteor}, \mbox{chrF}~\cite{popovic-2015-chrf}, and even embedding-based \textsc{BERTScore}~\cite{bert_score} treat tables as flat text, often ignoring header alignment, units, or cell hierarchy.  \textsc{PARENT}~\cite{parent} partly grounds evaluation in the input source but still struggles with schema-level changes. Algorithmic and LLM-assisted metrics such as \textsc{H-Score} and \textsc{P-Score}~\cite{tang-etal-2024-struc} move toward structural sensitivity but differ in design: the former computes heuristic, rule-based structural and content similarity, while the latter leverages LLM judgments; both offer limited interpretability. \textsc{TabEval}~\cite{tabeval} improves semantic coverage by decomposing tables into atomic statements and applying textual entailment, yet incurs NLI overheads and often over-penalizes harmless layout differences.  The recent \textbf{\textsc{TabXEval}}~\cite{tabxeval} represents a step-change: its two-phase design \emph{TabAlign} for structural alignment and \emph{TabCompare} for semantic/syntactic checks delivers interpretable cell-level diagnostics and consistently balances \textit{sensitivity} and \textit{specificity}, achieving strong human correlation and placing it in the ``Goldilocks'' zone for robust evaluation.

\paragraph{Reference-less Evaluation and Remaining Gaps.}  
Metrics such as \textsc{QuestEval} and Data-\textsc{QuestEval}~\cite{dataquesteval} demonstrate that reference-less evaluation is viable by generating and answering questions over the source data, showing strong alignment with humans in data-to-text tasks.  
However, their reliance on generic QA signals often misses table-specific structural errors, unit inconsistencies, or localized discrepancies.  
Despite advances from overlap-based to LLM-driven and rubric-based methods, most existing approaches still emphasize either semantics or structure and condense diverse errors into a single opaque score, limiting error traceability and robustness under realistic perturbations.

\section{Conclusion and Future Work}

We introduced \textbf{\method{}}, a property-driven, reference-less framework for evaluating tabular generation through graph-based reasoning and interpretable, rubric-aware scoring. By unifying structured alignment, factual comparison, and sensitivity–specificity control within a single pipeline, \method{} delivers consistent, human-aligned judgments that remain robust under domain shifts and perturbation difficulty. Our accompanying benchmark, \textbf{\benchmark{}}, establishes a new standard for systematic stress testing of table metrics across six diverse domains and twelve controlled perturbation types.

Experiments demonstrate that \method{} not only correlates most strongly with human evaluations but also provides fine-grained, explainable diagnostics at both cell and table levels enabling actionable analysis of model and prompt behaviors. Beyond outperforming reference-based and LLM-judge baselines, it shows that reliable table evaluation is possible without explicit references by reasoning over grounded factual graphs.

Future work will focus on extending \method{} to richer structural formats such as hierarchical or multi-modal tables, and on distilling its LLM components into lightweight, domain-adaptive evaluators for scalable deployment. We envision \method{} as a foundation for \emph{trustworthy, interpretable evaluation} in structured generation supporting better model selection, alignment, and reward learning across real-world applications.

\section{Limitations}
\label{sec:limitations}

While \method{} achieves robust and interpretable evaluation, it has a few limitations. It relies on large language models for fact extraction and alignment, which adds computational cost and mild variability due to model stochasticity. The current implementation supports only structured digital tables (e.g., HTML, Markdown) and cannot yet handle tables embedded in images or PDFs requiring OCR or visual parsing. Finally, although \benchmark{} spans six diverse domains, it remains limited to English and synthetic perturbations, leaving real-world noise, multilingual data, and complex layouts for future exploration.

\section{Ethics Statement}
The authors affirm that this work adheres to the highest ethical standards in research and publication. Ethical considerations have been meticulously addressed to ensure responsible conduct and the fair application of computational linguistics methodologies. Our findings are aligned with experimental data, and while some degree of stochasticity is inherent in black-box Large Language Models (LLMs), we mitigate this variability by maintaining fixed parameters such as temperature, $top_p$, and $top_k$. Furthermore, our use of LLMs, including \texttt{GPT-5-nano}, \texttt{Gemma}, and \texttt{InternVL}, complies with their respective usage policies. To refine the clarity and grammatical accuracy of the text, AI based tools such as Grammarly and ChatGPT were employed. Additionally, human annotators who are also among the authors actively contributed to data labeling and verification, ensuring high-quality annotations. To the best of our knowledge, this study introduces no additional ethical risks.

\section*{Maintenance and Adoption Plan}

To support long-term use and community adoption, we release all artifacts under permissive terms and commit to active maintenance. The \method{} reference implementation (including \textit{Text2Graph}, \textit{Graph Alignment}, and \textit{Property-Driven Scoring}) is open-sourced at \url{https://github.com/CoRAL-ASU/TabReX} under the MIT License, and \benchmark{} is hosted on the Hugging Face Hub with versioned revisions so that published numbers remain reproducible even as the dataset grows.

Ongoing maintenance is coordinated by \textbf{Junha Park} (\texttt{jpark284@asu.edu}) as the primary point of contact, with bug reports, feature requests, and new-domain contributions handled through GitHub Issues and Pull Requests. We will publish tagged releases for breaking changes and track upstream LLM judge deprecations (e.g., swapping \texttt{gpt-5-nano} for successor models) so that \method{} continues to function as provider APIs evolve. Contributions of additional evaluation domains, perturbation types, and alternative judge back-ends (including the embedding-based matcher described in Appendix~\ref{supsec:kg_validation}) are explicitly welcomed via the contribution guidelines in the repository.

\section*{Acknowledgments}
We thank the Complex Data Analysis and Reasoning Lab at Arizona State University for computational support. We also thank Adobe Research for supporting this work. Finally, we like to thank the anonymous reviewers for valuable feedbacks which helped improving the manuscript.

\bibliography{custom}

\appendix

\section*{Appendix}

\section{Human Evaluation Protocol}
\label{supsec:human_eval}

Human annotators were instructed to evaluate the similarity of generated tables to the gold (ground-truth) tables whenever available or against source text following a consistent rubric. Each annotation batch contained one gold table and five generated candidates. Annotators ranked candidates from 1 (best) to 5/12 (depending on task) (worst) based on their structural and contextual fidelity to the gold table.

\paragraph{Structural Factors.}  
Annotators prioritized structural integrity in the following order:  
(1) \textit{Column Missing} - tables omitting columns were penalized most heavily;  
(2) \textit{Column Extra} - extra columns ranked lower in case of ties;  
(3) \textit{Row Missing} and (4) \textit{Row Extra} - missing or spurious rows reduced rank;  
(5) \textit{Cell Missing} and (6) \textit{Cell Extra} - missing or redundant cells influenced ranking proportionally;  
(7) \textit{Partial Mismatching Severity} - deviations in value accuracy or format were also considered.

\paragraph{Contextual Factors.}  
Within equal structural quality, contextual accuracy guided ranking:  
(1) string-value mismatches,  
(2) numeric, boolean, or date-time inaccuracies,  
(3) inconsistencies in list-type entries, and  
(4) deviations in other less common data types.

\paragraph{Tie-Breaking.}  
In case of ties, rankings were determined by the number of affected cells within rows and columns. Column headers with semantically incorrect or mismatched meanings were treated as “wrong columns” and penalized equivalently to missing columns.

This rubric ensured consistent and interpretable human rankings aligned with the metric’s property-driven principles.

\section{Walk-Through Example of \method}
\label{supsec:walkthrough}

For full details of the formalism, please refer to the main paper.
Here we provide only the default hyperparameters and a worked example to show how
the score is computed in the reference-less setting.

\paragraph{Hyperparameters.}
\begin{table}[ht]
\centering
\resizebox{\linewidth}{!}{
\begin{tabular}{llc}
\toprule[1.5pt]
\textbf{Symbol} & \textbf{Meaning} & \textbf{Value} \\
\midrule
$\beta_{\mathrm{MI}}$ & Weight for Missing Information (MI) & 1.0 \\
$\beta_{\mathrm{EI}}$ & Weight for Extra Information (EI) & 0.9 \\
$\beta_{\mathrm{partial}}$ & Weight for partially correct cell values & 0.8 \\
$\alpha_{r}$ & Row-level (subject) structural weight & 0.9 \\
$\alpha_{c}$ & Column-level (predicate) structural weight & 1.0 \\
$\alpha_{\mathrm{cell}}$ & Cell-level (object) structural weight & 0.8 \\
$\omega_{p}$ & Scaling factor for partial deviation $\gamma$ & 0.9 \\
\bottomrule[1.5pt]
\end{tabular}}
\caption{Default \method{} hyperparameters.}
\label{tab:hyperparams}
\end{table}

\paragraph{Setup.}
Let $\mathcal{G}_{S}$ be the source-text evidence graph
and $\mathcal{G}_{T}$ the generated-table graph.
All counts below are measured relative to $\mathcal{G}_{S}$.
Assume
\[
N_{r}=5,\quad N_{c}=4,\quad N_{\mathrm{cell}}=20 ,
\]
with discrepancies:
\[
\mathrm{MI}_{r}=1,\quad
\mathrm{EI}_{c}=1,\quad
\mathrm{MI}_{\mathrm{cell}}=2,\quad
\mathrm{EI}_{\mathrm{cell}}=1 ,
\]
and two partially aligned cells with normalized deviations $0.2$ and $0.5$.

\paragraph{Step 1: Table-level penalty.}
\begin{align*}
\text{TablePenalty}
&= \beta_{\mathrm{MI}}\alpha_{r}\tfrac{\mathrm{MI}_{r}}{N_{r}}
 + \beta_{\mathrm{MI}}\alpha_{c}\tfrac{\mathrm{MI}_{c}}{N_{c}} \notag\\
&\quad+ \beta_{\mathrm{EI}}\alpha_{r}\tfrac{\mathrm{EI}_{r}}{N_{r}}
 + \beta_{\mathrm{EI}}\alpha_{c}\tfrac{\mathrm{EI}_{c}}{N_{c}} \notag\\
&= 1.0(0.9\tfrac{1}{5})
 + 0.9(1.0\tfrac{1}{4}) \notag\\
&= 0.18 + 0.225 = 0.405 .
\end{align*}

\paragraph{Step 2: Cell-level penalty.}
Partial-match deviations:
\begin{align*}
\gamma_{1}&=\omega_{p}\cdot 0.2=0.18, &
\gamma_{2}&=\omega_{p}\cdot 0.5=0.45, \\
\sum\nolimits_{i}\gamma_{i}&=0.63 .
\end{align*}

\begin{align*}
\text{CellPenalty}
&= \beta_{\mathrm{MI}}\alpha_{\mathrm{cell}}\tfrac{\mathrm{MI}_{\mathrm{cell}}}{N_{\mathrm{cell}}}
 + \beta_{\mathrm{EI}}\alpha_{\mathrm{cell}}\tfrac{\mathrm{EI}_{\mathrm{cell}}}{N_{\mathrm{cell}}} \notag\\
&\quad+ \beta_{\mathrm{partial}}\alpha_{\mathrm{cell}}
        \tfrac{1}{N_{\mathrm{cell}}}\sum\nolimits_{i}\gamma_{i} \notag\\
&= 1.0\!\times\!0.8\!\times\!\tfrac{2}{20}
 + 0.9\!\times\!0.8\!\times\!\tfrac{1}{20} \notag \\
 & + 0.8\!\times\!0.8\!\times\!\tfrac{0.63}{20} \notag\\
&= 0.08 + 0.036 + 0.0202
= 0.1362 .
\end{align*}

\paragraph{Step 3: Final score.}
\begin{align*}
\mathcal{S}_{\text{\method}}
&= \text{TablePenalty} + \text{CellPenalty} \notag\\
&= 0.405 + 0.1362
= 0.5412 .
\end{align*}

\paragraph{Interpretation.}
The example shows that both structural discrepancies
(missing rows, extra columns) and factual deviations
(partially mismatched cell values) jointly contribute
to the final reference-less \method\ score.

\section{Validation of LLM-based KG Extraction}
\label{supsec:kg_validation}

A natural concern with \method{} is that the summary-KG extractor is itself an LLM: if the extractor is unreliable, every downstream score inherits that noise. We address this concern with an empirical study on a 30-sample stratified validation set (6 domains $\times$ 5 summaries per domain, sampled uniformly at random).

\paragraph{Gold KG Construction.}
Human annotators manually authored the gold KG for each of the 30 summaries, producing \textbf{364 triples} in total (mean $12.1$, range $6$$23$). Triples follow a \texttt{(subject, predicate, object)} schema.

\paragraph{Extraction Quality.}
We compare three extractors against the gold KG: (i) \textbf{gpt-5-nano}, (ii) \textbf{Qwen3.5-27B}, and (iii) \textbf{Gemma-4-26B}. We report micro-averaged triple P/R/F1 under four matchers of increasing tolerance:
\begin{itemize}[leftmargin=*,topsep=2pt,itemsep=1pt]
    \item \textbf{Exact} - normalized string equality on all three positions.
    \item \textbf{Cross} - token-bag Jaccard $\geq 0.6$ allowing components to slide across subject/predicate/object positions.
    \item \textbf{Embed} - whole-triple cosine on \texttt{"s\,|\,p\,|\,o"} using \texttt{BAAI/bge-base-en-v1.5} (L2-normalized), with greedy 1-to-1 matching at threshold $0.85$. This matcher is \emph{LLM-free} and therefore independent of any judge model.
    \item \textbf{Semantic} - LLM-judge alignment using the same Graph Alignment prompt (\promptref{prompt:alignment}) employed by \method{}.
\end{itemize}

\begin{table}[ht]
\centering
\resizebox{\linewidth}{!}{
\begin{tabular}{lcccc}
\toprule[1.5pt]
\textbf{Model} & \textbf{Exact F1} & \textbf{Cross F1} & \textbf{Embed F1} & \textbf{Semantic F1} \\
\midrule
gpt-5-nano       & 0.262 & 0.450 & \textbf{0.854} & \textbf{0.914} \\
Qwen3.5-27B      & 0.242 & 0.450 & \textbf{0.903} & \textbf{0.869} \\
Gemma-4-26B-A4B  & 0.261 & 0.361 & \textbf{0.810} & \textbf{0.840} \\
\bottomrule[1.5pt]
\end{tabular}}
\caption{Triple-level P/R/F1 of three LLM extractors against the gold KG under four matchers of increasing tolerance.}
\label{tab:kg_extraction_f1}
\end{table}

Exact and cross F1 are low because the extractors paraphrase entities and predicates freely (e.g.\ \emph{``has\_value''} vs.\ \emph{``recorded\_value''}; \emph{``Q2 2023''} vs.\ \emph{``second quarter of 2023''})-exactly the kind of variation \method's alignment stage is designed to absorb. Under the two paraphrase-tolerant matchers, all three extractors recover $0.81$$0.91$ F1, and the two matchers agree within $0.06$ F1 on every model despite being built on independent signals (LLM judgment vs.\ dense embeddings). This cross-check rules out the circularity risk that a single LLM judge might over-credit its own family's paraphrases.

\begin{table}[ht]
\centering
\resizebox{0.85\linewidth}{!}{
\begin{tabular}{lcc}
\toprule[1.5pt]
\textbf{Model} & \textbf{Entity Coverage} & \textbf{Faithfulness} \\
\midrule
gpt-5-nano       & 0.794 & 0.884 \\
Qwen3.5-27B      & 0.866 & 0.892 \\
Gemma-4-26B  & 0.808 & 0.922 \\
\bottomrule[1.5pt]
\end{tabular}}
\caption{Auxiliary per-sample metrics (macro-averaged over 30 samples). \textbf{Entity coverage} is the fraction of unique gold subjects whose tokens are recovered (by $\geq 80\%$ token overlap) in the extractor's subjects or predicates. \textbf{Faithfulness} is the fraction of extracted-triple objects whose numeric values (or all non-trivial tokens) also appear in the source summary, measuring whether the extractor hallucinates values not grounded in the input.}
\label{tab:kg_aux}
\end{table}

\paragraph{Does Extractor Choice Move the Final \method{} Score?}
This is the bottom-line check. For each of the 30 samples we run the full \method{} pipeline \textbf{four times}, varying only the summary-KG source: gold KG, gpt-5-nano KG, Qwen3.5-27B KG, and Gemma-4-26B KG. The table graph and scoring rule are fixed; alignments are cached by a hash of \texttt{(judge, summary-KG, table-KG)}.

\begin{table}[ht]
\centering
\resizebox{\linewidth}{!}{
\begin{tabular}{lccc}
\toprule[1.5pt]
\textbf{LLM-extracted KG} & \textbf{Spearman $\rho$} & \textbf{Kendall $\tau$} & \textbf{$p$-value ($\rho$)} \\
\midrule
gpt-5-nano       & \textbf{+0.822} & +0.689 & $2.5 \times 10^{-8}$ \\
Qwen3.5-27B      & \textbf{+0.711} & +0.571 & $1.1 \times 10^{-5}$ \\
Gemma-4-26B  & \textbf{+0.830} & +0.666 & $1.4 \times 10^{-8}$ \\
\bottomrule[1.5pt]
\end{tabular}}
\caption{Rank correlation between \method{} scores computed with the gold summary KG and with each LLM-extracted summary KG across the 30 validation samples.}
\label{tab:kg_correlation}
\end{table}

Mean \method{} scores differ from gold by less than $1\%$ (gold $39.71$; extractors $39.43 / 40.04 / 39.98$). Rank correlations are strong and highly significant for all three extractors, including two open-weight models from different families. The final \method{} score is therefore stable under extractor choice-the LLM extractor is a well-controlled component rather than a metric threat.

\section{Prompt Templates}
\label{supsec:prompts}
\mytcbinputwide{prompts/Tab2Text.tex}{Table Summary Generation}{0}{bw:domain}{prompt:tab2text}
\mytcbinputwide{prompts/planning.tex}{Perturbation Planning}{0}{bw:domain}{prompt:planning}
\mytcbinputwide{prompts/T2G.tex}{Text2Graph}{0}{bw:domain}{prompt:t2g}
\mytcbinputwide{prompts/alignment.tex}{Graph Alignment}{0}{bw:domain}{prompt:alignment}

\end{document}